\documentclass[11pt,titlepage]{article}
\usepackage[margin=1.2in]{geometry}
\usepackage{amsmath, amssymb, amsbsy, amsthm, array, mathrsfs}
\usepackage{physics}
\usepackage{amssymb,latexsym,verbatim}
\usepackage{color}
\usepackage{dsfont, bbm}

\newcommand{\R}{\mathbb{R}}

\newcommand{\noin}{\noindent}
\newcommand{\bee}{\begin{eqnarray*}}
\newcommand{\ene}{\end{eqnarray*}}
\newcommand{\bec}{\begin{center}}
\newcommand{\enc}{\end{center}}
\newcommand{\be}{\begin{equation}}
\newcommand{\ee}{\end{equation}}

\newcommand{\ep}{\varepsilon}
\newcommand{\mb}{\mathbf}
\newcommand{\bs}{\boldsymbol}
\newcommand{\tb}{\textbf}
\newcommand{\pend}{$\square$}
\newcommand{\vs}{\vskip 3mm}
\newcommand{\bi}{\begin{itemize}}
\newcommand{\ei}{\end{itemize}}

\begin{document}
\title{\LARGE 
Non-asymptotic analysis of the performance of the penalized least trimmed squares in sparse models
} 
\vs
\vs

\author{ {\sc 
Yijun Zuo}\\[3ex]
         {\small Department of Statistics and Probability} \\[2ex]
         {\small Michigan State University, East Lansing, MI 48824} \\[2ex]
         {\small 
         zuo@msu.edu}\\[6ex]
     }
\date{\today}

\maketitle
\vskip 3mm
{\small

\begin{abstract}
The least trimmed squares (LTS)  estimator is a renowned robust alternative to the classic 
least squares estimator and is
 popular in location, regression, machine learning, and AI literature.	Many studies exist on LTS, including its robustness, computation algorithms, extension to non-linear cases,
asymptotics, etc. The LTS has been applied in the penalized regression in a high-dimensional real-data sparse-model setting where dimension $p$ (in thousands) is much larger than sample size $n$ (in tens, or hundreds). 
In such a practical setting, the sample size $n$ often is the count of sub-population that has a special attribute (e.g. the count of patients of Alzheimer's, or Parkinson's, Leukemia, or ALS, etc.) among a population with a finite fixed size N. Asymptotic analysis 
assuming that $n$ tends to infinity is not practically convincing  and legitimate
 in such a scenario. 
A non-asymptotic or finite sample analysis will be more desirable and feasible. 
\vs
\noin
This article establishes some finite sample (non-asymptotic) error bounds for estimating and predicting based on LTS with high probability for the first time.

\bigskip

\bigskip
\par

\noindent{\bf Key words and phrase:}  penalized least trimmed squares estimators, spare regression model, existence and uniqueness,  
finite sample prediction error bound,  finite sample estimation error bound.
\bigskip
\par
\noindent {\bf Running title:} Non-asymptotic analysis  of the performance of the least trimmed squares.
\end{abstract}
}

\setcounter{page}{1}

\section {Introduction}

\subsection{The least trimmed squares and its penalized versions}
In  classical multiple linear regression analysis, it is assumed that there is a relationship for a given data set $\{(\bs{x}^{\top}_i, y_i)^{\top}: i\in \{1,\cdots, n\}\}$: 
\be
y_i=(1,\bs{x}^{\top}_i)\bs{\beta}_0+{e}_i:=\bs{v}^{\top}_i\bs{\beta}_0+e_i,~~ i\in \{1,\cdots, n\},  \label{model.eqn}
\ee
where {$e_i$ (an error term, a random variable, and is assumed to have a zero mean and unknown variance $\sigma^2$ in the classic regression theory) and $y_i$}  are in $\R^1$, ${\top}$ stands for the transpose, $\bs{\beta}_0=(\beta_{01}, \cdots, \beta_{0p})^{\top}$, the true unknown parameter, {and}~ $\bs{x_i}=(x_{i1},\cdots, x_{i(p-1)})^{\top}$ is in $\R^{p-1}$ ($p\geq 2$) and could be random.  It is seen that $\beta_{01}$ is the intercept term.  
 The classic assumptions such as linearity and homoscedasticity are implicitly assumed here. 
\vs
The goal is to estimate the $\bs{\beta}_0$ via the given sample $\mb{z}^{(n)}
:=\{(\bs{x}^{\top}_i, y_i)^{\top}: i\in \{1,\cdots, n\}\}$. (hereafter it is implicitly assumed that they are independent, identically distributed from parent $(\bs{x}^\top, y)$).
For a candidate coefficient vector $\bs{\beta}$, call the difference between $y_i$ (observed) and $\bs{w^{\top}_i}{\bs{\beta}}$ (predicted),
 the ith residual, $r_i(\bs{\beta})$, ($\bs{\beta}$ is often suppressed). That is,\vspace*{-2mm}
\be r_i:= {r}_i(\bs{\beta})=y_i-(1, \bs{x}^\top_i) \bs{\beta}_0=y_i-\bs{v^{\top}_i}{\bs{\beta}}.\label{residual.eqn}
\ee
To estimate $\bs{\beta}_0$, the classic \emph{least squares} (LS) minimizes  the sum of squares of residuals (SSR),
$$\widehat{\bs{\beta}}_{ls}:=\arg\min_{\bs{\beta}\in\R^p} \sum_{i=1}^n r^2_i. $$
Alternatively, one can replace the square above with the absolute value to obtain the least absolute deviations estimator (aka, $L_1$ estimator, in contrast to the $L_2$ (LS) estimator).
\vs
 Due to its great computability and optimal properties when the error $e_i$ follows a Gaussian 
  distribution, the LS estimator is popular in practice across multiple disciplines.
It, however, can behave badly when the error distribution departs from the Gaussian assumption,
particularly when the errors are heavy-tailed or contain outliers.
In fact, both $L_1$ and $L_2$  estimators have the worst $0\%$ asymptotic breakdown point, in sharp contrast to the $50\%$ of the least
trimmed squares (LTS) estimator (Rousseeuw (1984) (R84), and Leroy (1987) (RL87)). 
 The latter is one of the most robust alternatives to the LS estimator.
LTS is popular in the literature because of its fast computability and high robustness and often serves as the initial estimator for many high breakdown point iterative procedures (e.g., S- (Rousseeuw and Yohai (1984)) and MM- (Yohai (1987)) estimators). The LS utilizes all n squared residuals. Let $h$ be the count of squared residuals that will be utilized,
the LTS is defined as
the  minimizer of the sum of $h$  {smallest squares} of residuals. Namely, \vspace*{-2mm}
\be
\widehat{\bs{\beta}}_{lts}:=\arg\min_{\bs{\beta}\in \R^p} \sum_{i=1}^h r^2_{i:n},\label{lts.eqn}
\ee
\vspace*{-.5mm}

\noin
where $r^2_{1:n}\leq r^2_{2:n}\leq \cdots\leq r^2_{n:n}$ are the ordered squared residuals and $\lceil n/2\rceil\leq h < n$ and $\lceil x \rceil$ is the ceiling function.
\vs

There are copious studies on the LTS in the literature.  
Most focused on its computation, see, e.g., RL87, 
 Stromberg (1993), 
  H\"{o}ssjer (1995), 
  Rousseeuw and Van Driessen (1999, 2006), 
  Agull\'{o} (2001), 
  Hofmann et al. (2010), 
 and Klouda (2015). 
Others 
addressed the asymptotics, see, e.g., RL87, and  Ma\v{s}\'{i}\v{c}ek (2004) 
 studied the asymptotic normality of the LTS, but limited to the location case, that is, when $p=1$.
 V\'{i}\v{s}{e}k (2006a b c) 
addressed the asymptotics of the LTS 
 without employing advanced technical tools in a series (three) of articles for the consistency, root-n consistency, and
asymptotic normality, respectively. The analysis is technically demanding 
and with difficult verified 
assumptions $\mathcal{A, B, C}$ and limited to the non-random vectors $\bs{x}_i$s case.
Zuo (2024) provided a study of asymptotics of LTS
without those assumptions and limitations.
\vs 
The LTS has been
extended to 
penalized regression setting with a sparse model where dimension $p$ (in thousands) much larger than sample size $n$ (in tens, or hundreds), see, e.g.,
 Alfons et al. (2013), and Kurnaz et al. (2018). Here is 
  a more general definition.\vs\noin
 \tb{Definition 1.1 Penalized LTS regression estimator}. One wants to
 minimize the objective function
 in (\ref{lts.eqn}) subject to two constraints: $\ell_{\gamma}$-constraint $\sum_{i=1}^p|\beta_i|^{\gamma}\leq t_1$, $t_1\geq 0$, $\gamma\geq 1$; and $\ell_2$-constraint $\sum_{i=1}^p\beta^2_i \leq t_2$, $t_2\geq 0$,
 the minimizer  is
 \be
\widehat{\bs{\beta}}^n_{lts-enet}(h,\lambda_1,\lambda_2, \gamma)=\arg\min_{\bs{\beta}\in \R^p}\Big\{\frac{1}{n} \sum_{i=1}^n (w_ir_i^2)+ \lambda_1\sum_{i=1}^p |{\beta}_i|^{\gamma}+\lambda_2 \|\bs{\beta}\|^2_2 \Big\}, \label{lts-en.eqn}
\ee
where $\lambda_i:=\lambda(t_i)\geq 0$, and $w_i:=w_i(\bs{\beta}):=w_i(\bs{\beta}, h, r_i, \bs{z}^{(n)})=\mathds{1}\big( r_i^2 \leq r^2_{(i:h)}  \big)$,   $\|\bs{x}\|_q=(\sum_{i=1}^n x_i^q)^{1/q}$ is the $\ell_q$-norm for vector $\bs{x} \in \R^n$. 
When $\lambda_1=\lambda_2=0$,  $\widehat{\bs{\beta}}^n_{lts-enet}(h,\lambda_1,\lambda_2,\gamma)$ becomes the regular $\widehat{\bs{\beta}}_{lts}$ in (\ref{lts.eqn}). When $\lambda_2=0$, it recovers the $\widehat{\bs{\beta}}_{lts-lasso}$ in   Alfons et al. (2013) for a fixed $h$. Hereafter, for simplicity write $\widehat{\bs{\beta}}^n_{lts-enet}$ for
 $\widehat{\bs{\beta}}^n_{lts-enet}(h,\lambda_1,\lambda_2,\gamma)$.\vs

\subsection{Non-asymptotic analysis}
In the small sample size $n$ (in tens, or hundreds) and large dimension $p$ (in thousands) real data set scenario, the usual asymptotic analysis has all theoretical merits but falls short in practical use. A non-asymptotic analysis is more desirable. 
Here one wants  to bound  with a high probability the estimation error $\|\widehat{\bs{\beta}}^n_{lts-enet}-\bs{\beta}_0\|^2_2$ or prediction error $\|\bs{X}\big(\widehat{\bs{\beta}}^n_{lts-enet}-\bs{\beta}_0\big)\|^2_2$, where 
 $\bs{X}=(\bs{v}_1,\cdots, \bs{v}_n)^\top$ and $\bs{v}_i$ is defined in (\ref{model.eqn}).
\vs \noin
The rest of the article is organized as follows. Section 2 first establishes the existence and uniqueness of the minimizer in the RHS of (\ref{lts-en.eqn}). Section 3 will establish
the finite sample error bounds for the estimation and prediction of $\widehat{\bs{\beta}}^n_{lts-enet}$. Concluding remarks in Section 4 ends the article. Proofs of main results are deferred to an Appendix.

\section{Existence and Uniqueness} \label{sec.2}\vs

Existence and uniqueness are implicitly assumed for many other penalized regression estimators in the literature. They are preliminary conditions for any further study of the property of the estimators. We formally address them for $\widehat{\bs{\beta}}^n_{lts-enet}$  below.
\vs
\noindent
\tb{Theorem 2.1} For a given sample $\bs{z}^{(n)}$, and $\lambda_i\geq0$, a fixed $h$, $\widehat{\bs{\beta}}^n_{lts-enet}(h,\lambda_1,\lambda_2,\gamma)$ in (\ref{lts-en.eqn}) \vs \noin
(i) always exists; furthermore, (ii) it is unique.
\vs

\noindent
\tb{Proof}: Existence is based on the argument of continuity and compactness; uniqueness is based on strict convexity, for details see the Appendix. \hfill \pend
\vs \noin
The proof of the above theorem needs the following result.
\vs
 \noindent
 \tb{Lemma 2.2} Let $S\subset \R^p$ be an open set and $f(\bs{x})$: $\R^p \to \R^1$ be  strictly convex over $S$ and continuous over $\overline{S}$ (the closure of $S$). Let $\bs{x}^*$ be the global minimum of $f(\bs{x})$ over $S$ and $\bs{y}$ be a point on the boundary of $S$,  then $f(\bs{y}) > f(\bs{x}^*)$.
 \vs
 \noin
 \tb{Proof:} See the Appendix. \hfill \pend
 \vs

\section{Finite sample prediction and estimation error bounds \label{sec.3}}
\subsection{Prediction error bound}
\noin
In this section we assume that the true model is $\bs{Y}=\bs{X}\bs{\beta}_0+\bs{e}$ where $\bs{Y}=(y_i,\cdots, y_n)^\top$, $\bs{X}:=\bs{X}_{n \times p}=(\bs{v}_1,\cdots, \bs{v}_n)^\top$, and $\bs{e}=(e_1,\cdots, e_n)^\top$ with $y_i$, $e_i$, and $\bs{v}_i=(1,\bs{x}^\top_i)^\top$ defined in (\ref{model.eqn}) and  (\ref{residual.eqn}).
  We investigate the difference between $\bs{X}\widehat{\bs{\beta}}^n_{lts-enet}$ and $\bs{X}\bs{\beta}_0$ (prediction error). Write $\widehat{\bs{\beta}}^n$ for $\widehat{\bs{\beta}}^n_{lts-enet}$ for simplicity. 
\vs
In light of $w_i$ in (\ref{lts-en.eqn}), for a given sample $\bs{z}^{(n)}$ and any $\bs{\beta}$, introducing an index set $I(\bs{\beta})$  as 
\be I(\bs{\beta}):=\{i: w_i=1, i\in\{1,2,\ldots, n\}\}. \label{index.eqn}\ee   
With $w_i\in \{0, 1\}$ in (\ref{lts-en.eqn}) as diagonal entries of a diagonal matrix, 
let $\mbox{diag}(w_1, \cdots, w_n):={D}(\bs{\beta})$. Let ${A}$ be an $n$ by $n$ symmetric positive semidefinite matrix, a norm (or more precisely seminorm) induced by ${A}$ is $\|\bs{x}\|^2_{{A}}=\bs{x}^\top{A}\bs{x}$ for any $\bs{x}\in \R^n$. Although $\widehat{\bs{\beta}}^n$ can provide predictions for all $i$, but we just employed residuals $r_i$ with $i\in I(\widehat{\bs{\beta}}^n)$ in (\ref{lts-en.eqn}), so instead of
  looking at $\|\bs{X}\big(\widehat{\bs{\beta}}^n-\bs{\beta}_0\big)\|^2$, we will focus on  $\|\bs{X}\big(\widehat{\bs{\beta}}^n-\bs{\beta}_0\big)\|^2_{D(\widehat{\bs{\beta}}^n)}$, the squared prediction error.
\vs
Let $M$ be an n by n zero matrix. Sample $h$ of its diagonal entries and replace them by one, denote the resulting matrix by $D_1$ and repeat the sampling process until exhaust all 
$L= \binom{n}{h}$ 
possible combinations. 
Call the collection of all distinct $D_i$, $ i \in \{1,\cdots, L\}$ as 
$\mathbb{D}$. Clearly $|\mathbb{D}|=L$ and $D(\widehat{\bs{\beta}}^n)\in \mathbb{D}$ for any given $\bs{z}^{(n)}$ and
$\widehat{\bs{\beta}}^n$. Define $\Theta$ to be a compact set of $\bs{\beta}$s such that $\bs{\beta}_0 \in \Theta$ and $\bs{\beta} \in \Theta$ if $\|\bs{\beta}\|^{\gamma}_{\gamma}\leq t_1$, where $t_1$ is given in the Definition 1.1.
Partition the sample space $\Omega$ in the probability triple $(\Omega, \mathcal{F}, \mathcal{P})$ into disjoint pieces $\Omega_i$, $i=1,\cdots, L$: 
\be
\Omega_i:=\{\omega\in \Omega: D\big(\widehat{\bs{\beta}}^n (\omega)\big)=D_i \} \label{omega_i.eqn}
\ee
\vs
\noindent
\tb{Lemma 3.1} Assume that $\bs{\beta}_0$ is the true parameter of the model in (\ref{model.eqn}), $\widehat{\bs{\beta}}^n:= \widehat{\bs{\beta}}^n_{lst-enet}$  defined in (\ref{lts-en.eqn}). We have
\begin{align}
\frac{1}{n}\|\bs{X}\big(\widehat{\bs{\beta}}^n-\bs{\beta}_0\big)\|^2_{{D}(\widehat{\bs{\beta}}^n)}&\leq 
 \frac{2}{n} \bs{e}^\top {D}(\widehat{\bs{\beta}}^n)\bs{X}(\widehat{\bs{\beta}}^n-\bs{\beta}_0)
+\frac{1}{n}\big(\|\bs{e}\|^2_{{D}(\bs{\beta}_0)}-\|\bs{e}\|^2_{{D}(\widehat{\bs{\beta}}^n)}\big) \nonumber\\[1ex]
&+\lambda_1\|\bs{\beta}_0\|^{\gamma}_{\gamma}+\lambda_2\|\bs{\beta}_0\|^2_2-\lambda_1\|\widehat{\bs{\beta}}^n\|^{\gamma}_{\gamma}-\lambda_2\|\widehat{\bs{\beta}}^n\|^2_2. \label{base-inequality.eqn}
\end{align}

\noindent
\tb{Proof}: Two facts are essential for the proof, one fact is the true model: $\bs{Y}=\bs{X}\bs{\beta}_0+\bs{e}$, and the second one is $\hat{\bs{\beta}}^n$ is the minimizer of the RHS of (\ref{lts-en.eqn}), for details 
see the Appendix.
\hfill \pend
\vs
Write $(\bs{e}^*)^\top:=(e^*_1,\cdots, e^*_n)$ with $e^*_i=e_i\mathds{1}\big(i\in I(\widehat{\bs{\beta}}^n)\big)$. Then
$\bs{e}^\top{D}(\widehat{\bs{\beta}}^n)=(\bs{e}^*)^\top$  
is equivalent to keep $h$ components of $\bs{e}$ unchanged and replace other components by zero.
 Define, for any $D\in \mathbb{D}$ and any $\bs{\beta}\in \Theta$, the sets
$$
\mathscr{S}_1: = \left\{\max_{1\leq j\leq p}2|\bs{e}^\top D\bs{x}^{(j)}|/n\leq q_1 \right\}, ~~
\mathscr{S}_2: = \left\{\|\bs{e}\|^2_{D(\bs{\beta}_0)}/\sigma^2-h\leq q_2\right\}.$$
$$
\mathscr{S}_3: = \left\{h-\|\bs{e}\|^2_{D(\bs{\beta})}/\sigma^2\leq q_3\right\}.
$$
where 
$\bs{x}^{(j)}$ is the $j$th column of the fixed  design matrix $\bs{X}_{n \times p}$. $\sigma^2$ is the variance of $e_i$, $q_i$ defined below.
As in the literature (see, e.g., page 20 of Pauwels (2020)), assume that $\max_{1\leq j\leq p}\|\bs{x}^{(j)}\|_2/\sqrt{n}\leq 1$. 
\vs
In the classical setting, { it is assumed that \tb{A}:
$e_i$ in (\ref{model.eqn}) are i.i.d. $N(0,\sigma^2)$.} It is needed for the second result below, but for the first, it can be relaxed {\tb{A0}: $e_i$ in (\ref{model.eqn}) are i.i.d. sub-Gaussian variables with variance proxy $\sigma^2$.} For the definition of the latter, we refer to  Definition 1.2 of Rigollet and H\"{u}tter (2017)
and/or  Theorem 2.1.1 of Pauwels (2020). 
\vs
\noin
\tb{Lemma 3.2}  For any $D\in \mathbb{D}$,  { (i) assume that \tb{A0} holds,}
then $(\bs{e})^\top D \bs{x}^{(j)}/\sqrt{n}$ is a sub-Gaussian variable with variance proxy $\sigma^2$.
(ii) {assume that \tb{A} holds,}, then 
$\|\bs{e}\|^2_{D}/\sigma^2 $ 
follow a ${\chi}^2$ distribution with $h$ degrees of freedom. 
(iii) (i) and (ii) hold uniformly over $\mathbb{D}$ and  $\Theta$ with
$\bs{\beta} \in \Theta$ and $D=D(\bs{\beta})\in \mathbb{D}$.
\vs

\noindent
\tb{Proof}: Utilizing the equivalent definition of the sub-Gaussian variable given in Theorem 2.1.1 of Pauwels (2020), the desired result follows, see the Appendix.
\hfill \pend
\vs

\noin
\tb{Lemma 3.3} {Assume that \tb{A} holds}. For any $\delta\in (0,1)$ let $$q_1=2\sigma\sqrt{\frac{2(\log \frac{4pL}{\delta} )}{n}};~ 
q_2=2 \sqrt{\log(\frac{4L}{\delta})}\Big(\sqrt{h}+\sqrt{\log(\frac{4L}{\delta})}\Big); q_3=q_2-2\log(\frac{4L}{\delta}),$$
\noin
Then for any $D_i \in \mathbb{D}$, $i\in \{1, \cdots, L\}$ and
  $\Omega_i$,  set $p_i:=P(\Omega_i)$,  we have
\be
P(\mathscr{S}_1)\geq p_i-\delta/(2L);  ~~    
P(\mathscr{S}_2)\geq p_i-\delta/(4L);
P(\mathscr{S}_3)\geq p_i-\delta/(4L). \label{probility.ineq}
\ee
\vs
\noindent
\tb{Proof}: Employing the exponential upper bounds for the sub-Gaussian and Chi-square variables, the desired result follows. For details, see the Appendix. \hfill \pend

\vs
\noin
For simplicity, set $\gamma=1$ hereafter. Define for any $D \in \mathbb{D}$ and $\bs{\beta} \in \Theta$,
\begin{align}
\mathbb{G}(D, \bs{\beta}, \bs{e}; \bs{X}, \bs{\beta}_0, \lambda_1, \lambda_2):
&=\frac{2}{n} \bs{e}'{D}\bs{X}({\bs{\beta}}-\bs{\beta}_0)
+\frac{1}{n}\big(\|\bs{e}\|^2_{{D}(\bs{\beta}_0)}-\|\bs{e}\|^2_{{D}}\big) 
+g(\bs{\beta}, \bs{\beta}_0, \lambda_1, \lambda_2), \label{G-func.eqn}
\end{align}
where $g(\bs{\beta}, \bs{\beta}_0, \lambda_1, \lambda_2)=\lambda_1\|\bs{\beta}_0\|_{1}+\lambda_2\|\bs{\beta}_0\|^2_2-\lambda_1\|{\bs{\beta}}\|_{1}-\lambda_2\|{\bs{\beta}}\|^2_2$.\vs
\noin
\tb{Lemma 3.4} Select $\lambda_i$ such that
 $\lambda_1/2= q_1\geq \lambda_2$. {Assume that \tb{A} holds,} then for any $\delta\in (0,1)$, 
one has\vs \noin
(i) for any $\bs{\beta} \in \Theta$ with $D(\bs{\beta})=D_i\in \mathbb{D}$,  over $\Omega_i$ 
\begin{align*} 
P\Big( \omega \in \Omega_i, 
\bs{\beta} \in \Theta:
\mathbb{G}(D(\bs{\beta}), \bs{\beta}, \bs{e}; \bs{X}, \bs{\beta}_0, \lambda_1, \lambda_2) 
\leq \frac{\sigma^2}{n}\|\bs{\beta}_0\|_1C(\bs{\beta}_0,n, \sigma,q_1, q_2)\Big) & \\[2ex] 
\geq p_i-\delta/L,&
\end{align*}
 where $C(\bs{\beta}_0,n, \sigma,q_1, q_2):=2nq_1(2+\|\bs{\beta}_0\|_1)/\sigma^2+2 q_2/\|\bs{\beta}_0\|_1$;\vs\noin
(ii ) for any $\bs{\beta} \in \Theta$, over $\Omega$, one has
\begin{align*} 
P\Big( \mathbb{G}(D(\bs{\beta}), \bs{\beta}, \bs{e}; \bs{X}, \bs{\beta}_0, \lambda_1, \lambda_2) 
\leq \frac{\sigma^2}{n}\|\bs{\beta}_0\|_1C(\bs{\beta}_0,n, \sigma,q_1, q_2)\Big)&\\[2ex] \geq 1-\delta.&
\end{align*}
\vs
\noin
\tb{Proof:} In order to obtain a uniform result, we have to separate the proof into two steps, one is for the fixed case and the other is combining all possible fixed cases. For the details, see the Appendix. \hfill \pend
\vs
In light of all Lemmas above we are positioned to present the first main result.
\vs
\vs
\noindent
\tb{Theorem 3.1} Set $\gamma$ in (\ref{lts-en.eqn}) to be one and assume that \tb{A} holds, Set $\lambda_1/2= q_1\geq \lambda_2$. Then, for any $\delta\in (0,1)$, with probability at least $1-\delta$, one has
\be
\frac{1}{n}\|\bs{X}(\widehat{\bs{\beta}}^n-\bs{\beta}_0)\|^2_{D(\widehat{\bs{\beta}}^n)}\leq \frac{\sigma^2}{n}\|\bs{\beta}_0\|_1C(\bs{\beta}_0,n, \sigma,q_1, q_2).
 \label{error-bound.eqn}
\ee
\vs

\noindent
\tb{Proof}: See the Appendix. \hfill \pend
\vs
\noindent
\tb{Remarks 3.1}
\vs
\tb{(i)} 
 If we set $\lambda_2 = 0$, that is, just focus on the $\ell_1$ penalized estimation as the lasso, then the upper bound in (\ref{error-bound.eqn}) simplifies
to $2\lambda_1 \|\bs{\beta}_0\|_1 +\frac{2\sigma^2}{n} q_2$. We see that the rate of $\sqrt{\log p/n}$ convergence of the lasso is essentially recovered by the first term (see page 104 Corollary 6.1 of B\"{u}hlmann, P.  and Van De Geer(2011) (BVDG11)), provided that the order of $\log L$ is no higher than $\log p$ whereas the second term is the price one has to pay for the gain of robustness of $\widehat{\bs{\beta}}^n$ over the regular $\widehat{\bs{\beta}}^n_{lasso}$.
\vs

\tb{(ii)} In above discussions, we treat the unknown $\sigma$ as known. It appears in $q_1$ and in the upper bound of (\ref{error-bound.eqn}).
 In practice, we have to estimate it by an estimator, say $\widehat{\sigma}$ so that
$P(\widehat{\sigma}\geq\sigma)$ with high probability (say, $1-\delta/(3L)$, in this case, if we change $\delta/(2L)$,
$\delta/(4L)$ and $\log(4pL/\delta)$, $\log(4L)/\delta$ in Lemma 3.3 to $\delta/(3L)$, $\delta/(6L)$ and $\log(6pL/\delta)$,
$\log(6L/\delta)$, respectively,
 then Theorem 3.1 still holds). Such an estimator $\widehat{\sigma}$  has been given on page 104 of 
BVDG11.
\vs
\tb{(iii)} One limitation of Theorem 3.1 is that the design matrix is fixed. For the general random design $\bs{X}$ case, one can treat it following the approaches of 
Bartlett et al. (2012) and  
Gu\'{e}don et al. (2007).
\hfill \pend
\vs
\subsection{Estimation error bound}
Besides prediction error, one might also be interested in the estimation error, especially in the case of sparse setting ($p>n$ and there are only a few components of $\bs{\beta}_0$ being non-zero, for convenience we write $\bs{\beta}^0$ for $\bs{\beta}_0$ hereafter). Adopt the notations in the literature (see, e.g., p. 102 of BVDG11),  let 
\be
S_0=\{j: {\beta}^0_j\neq 0\} \label{S0.eqn}
\ee
be the index set of non-zero components of  $\bs{\beta}^0=(\beta^0_1,\cdots, \beta^0_p)^\top$. Denote the cardinality of $S_0$ by $s_0$, that is $s_0=|S_0|$. 
\vs

Let $S$ be an index set $S\subset \{1,2, \cdots, p\}$. For any $\bs{\beta}=(\beta_1,\cdots, \beta_p)^\top\in \R^p$,
define  $$\bs{\beta}_{S}= (\beta_1\mathds{1}(1\in S), \cdots, \beta_p\mathds{1}(p\in S))^\top,$$ 
Namely, a vector with its $j$ component being $\beta_j\mathds{1}(j\in S)$, $j\in \{1,2, \cdots, p\}$. It is obvious that $\bs{\beta}=\bs{\beta}_{S}+\bs{\beta}_{S^c}$ and $\bs{\beta}^\top_{S}\bs{\beta}_{S^c}=0$, where $A^c$ stands for the complement of the set $A$.
\vs
\noin
\tb{Lemma 3.5} {Assume that \tb{A} holds and let $\lambda_2=0$ and denote $\lambda_1$ by $\lambda$. For any $\delta\in (0,1)$ with probability at least $1-\delta$, one has
\be
\frac{2}{n}\|\bs{X}(\widehat{\bs{\beta}}^n-\bs{\beta}^0)\|^2_{D(\widehat{\bs{\beta}}^n)}
+\lambda\|\widehat{\bs{\beta}}^n_{S^c_0}\|_1\leq 3\lambda\|\widehat{\bs{\beta}}^n_{S_0}-\bs{\beta}^0_{S_0}\|_1+\eta,
\ee 
where $\eta:=4\hat{\sigma}^2{q_2}/{n}$ and $\hat{\sigma}$ is the one given in \tb{(ii)} of Remarks 3.1.

\vs
\noindent
\tb{Proof}: See the Appendix. \hfill \pend
\vs
\vs
\noin
\tb{Remarks 3.2}\vs
\tb{(i)}
The Lemma 3.5 implies that
\be
\|\widehat{\bs{\beta}}^n_{S^c_0}-\bs{\beta}^0_{S^c_0}\|_1\leq 3\|\widehat{\bs{\beta}}^n_{S_0}-\bs{\beta}^0_{S_0}\|_1+\eta/\lambda. \label{cone-condition}
\ee 
Without the $\eta$, (\ref{cone-condition}) is called cone condition in the literature, see e.g., Pauwels(2020) and Rigollet and H\"{u}tter (2017).\vs 
\tb{(ii)}
For simplicity write $\bs{X}_{*}$ for $D(\widehat{\bs{\beta}}^n)\bs{X}$. Define 
\be\mbox{MSE}(\bs{X}_{*}\widehat{\bs{\beta}}^n)=
\frac{1}{n}\|\bs{X}(\widehat{\bs{\beta}}^n-\bs{\beta}^0)\|^2_{D(\widehat{\bs{\beta}}^n)}=
(\widehat{\bs{\beta}}^n-\bs{\beta}^0)^\top\frac{\bs{X}^\top_{*}\bs{X}_{*}}{n} (\widehat{\bs{\beta}}^n-\bs{\beta}^0), \label{mse.eqn}
\ee
if $p<n$ and $A:=\frac{\bs{X}^\top_{*}\bs{X}_{*}}{n} $ has rank $p$, then one obtains the estimation error bound:
\be\|\widehat{\bs{\beta}}^n-\bs{\beta}^0\|^2_2\leq \frac{\mbox{MSE}(\bs{X}_{*}\widehat{\bs{\beta}}^n)}{\gamma_{min}(A)}, \label{estimation-bound-1}
\ee
where $\gamma_{min}(A)$ stands for the minimum eigenvalue of $A$.  Combining with Theorem 3.1, we obtain an upper bound for the estimation error. \hfill\pend\vs

\noin
Unfortunately, in a high-dimension setting, typically $p>n$, and the rank of $A$ must be less than $p$. So the above bound is invalid.  Denote a $p$ by $p$ identity matrix by $\mathbb{I}_{p \times p}$. 

\vs

\vs \noin
\tb{Definition 3.1~~Incoherence} (\cite{P20}\cite{RH17}) A matrix $\mathbb{X}\in \R^{n\times p}$ is said to have incoherence $k$ for some integer $k>0$ if
\be
\Big\|\frac{\mathbb{X}^\top\mathbb{X}}{n}-\mathbb{I}_{p \times p}\Big\|_{\infty}\leq \frac{1}{32k}, \label{inco.eqn}
\ee
where the $\|A\|_{\infty}$ denotes the largest element of $A$ in absolute value.
\vs
{
When $p\leq n$, it is often assumed in the machine learning literature that
\[{\mathbb{X}^\top\mathbb{X}}=n\mathbb{I}_{p \times p},\]
or equivalently, (i) $\frac{\|\bs{x}^{(i)}\|_2^2}{n}-1=0$, and (ii) $\frac{{\bs{x}^{(i)}}^{\top}\bs{x}^{(j)}}{n}=0$, where $\bs{x}^{(i)}$ is the ith column of matrix $\mathbb{X}$.
When $p>n$ we must modify the assumptions such as replacing the 0s above with a positive number such as $1/32k$
for some positive integer $k$, which leads to (\ref{inco.eqn}). For more discussions on  Incoherence, see pages 50, 59-60 of \cite{RH17}, page 26 of \cite{P20}, and pages 201-203 of \cite{W19}. }
\vs
Recall $\mathbb{D}$ defined in 
Section 3.1,
$D\in \mathbb{D}$ is a $n\times n$ diagonal matrix with all entries being zero except arbitrarily $h$ diagonal entries being one, obviously there are totally $L$ such matrices. Denote the collection of all such $D$s by $\mathbb{D}$ and $D\bs{X}$ by $\bs{X}_{*}$.\vs
\noin
\tb{Theorem 3.2}   {Assume that \tb{A} holds} and let $\lambda_2=0$ and denote $\lambda_1$ by $\lambda$. Assume that $\bs{X}_{*}$ has incoherence $k$ for some integer $k$ and any $D\in \mathbb{D}$ and $k\geq s_0\vee (p-s_0)/20$. Then, for any $\delta\in (0,1)$ with probability at least $1-\delta$, one has
\be
\|\widehat{\bs{\beta}}^n-\bs{\beta}^0\|^2_2\leq \frac{8}{3}\mbox{MSE}(\bs{X}_{*}\widehat{\bs{\beta}}^n)+\frac{1}{6k}\zeta^2,\label{estimation-bound}
\ee
where $\zeta:=\eta/\lambda$.

\vs
\noindent
\tb{Proof}: See the Appendix.
\hfill \pend

\section{Concluding remarks}
There are abundant discussions on the asymptotics of the penalized regression estimators in the sparse model settings where dimension $p$ (in thousands) is much larger than sample size $n$ (in tens, or hundreds). While most assuming $n$ goes to infinity, some authors fixed $p$, others allowed the $p$ to change with $n$ in certain rates. These studies have high mathematical merits but not so for practical significance. Finite sample studies are more desirable in real data scenarios.\vs
This article establishes estimation and prediction error bounds of penalized regression estimators based on the LTS for any
sample size $n$. These bounds have foundational importance for inference in real data scenarios, where sample size $n$ is fix and finite; while they also serve as reliable measures for the comparison of the performance of different penalized regression estimators in those real data scenarios.
\vs
Finite sample estimation and prediction error bounds of penalized regression estimators in the spare model setting are common in the literature when the objective function is convex, they are not so often when the objective function is non-convex (since it is much more harder to establish). For the non-convex LTS objective function, the bounds have never been addressed in the literature.\vs The results in this article, albeit being pioneers, have left rooms for further improvement. For exmple, to improve the upper bounds in Theorems 3.1 and 3.2 and obtain sharper ones for the estimation and prediction error of the LTS based penalized regression estimators.

\vspace{6pt}{}
%
\vs\vs \noin
\tb{Funding:}\vs \noin
{This author declares that there is no funding received for this study.}

\vs \noin
\tb{Acknowledgments}
{The author thanks Profs  Weng, H. and Shao, W. for
insightful comments and stimulating discussions; thanks also go to the two anonymous referees for their valuable comments and suggestions, all of which significantly improved the manuscript.}
\vs\noin
\tb{Conflicts of interest}
{This author declares that there is no conflict of interests/competing interests.}



%
%
\section[Appendix]{Proofs of main results}
\noindent
\tb{Proof of Theorem 2.1}
\vs
\noin
\tb{Proof}:
 $\lambda_1+\lambda_2=0$ case (that is the regular LTS case) has been treated in Zuo (2024) (Z24). 
 We treat $\lambda_1+\lambda_2>0$ case here.\vs
\tb(i) Denote the objective function on the RHS of (\ref{lts-en.eqn}) as
\be O_n(h,\bs{\beta},\lambda_1, \lambda_2, \gamma):=O(h,\bs{\beta}, \lambda_1, \lambda_2,   \gamma, \bs{z}^{(n)})=\frac{1}{n}\sum_{i=1}^{n}r_i^2w_i+\lambda_1 \sum_{j=1}^p|\beta_j|^{\gamma}+\lambda_2\sum_{i=1}^p\beta^2_i.\label{objective.eqn}
\ee
Name the three terms on the RHS above as $g(h, \bs{\beta}, \bs{z}^{(n)})$, $g_1(\lambda_1, \gamma, \bs{\beta})$, and $g_2(\lambda_2,\bs{\beta})$, respectively. It is readily seen that
the RHS of (\ref{lts-en.eqn}) is equivalent to minimizing $G(\bs{\beta}, h, \lambda_1, \gamma):=g(h, \bs{\beta}, \bs{Z}^{(n)})+g_1(\lambda_1, \gamma, \bs{\beta})$ subject to $\sum_{i=1}^p\beta^2_i \leq t_2$, $t_2\geq 0$, where $t_2$ defined in
Definition 1.1. 
\vs
By Lemma 2.2 of Z24,  
$g(h, \bs{\beta}, \bs{z}^{(n)})$ is continuous in $\bs{\beta}$ (this is not as obvious as one believed) while the continuity of $g_1(\lambda_1, \gamma,  \bs{\beta})$
in $\bs{\beta}$ is obvious. Therefore we have a continuous function of $\bs{\beta}$, $G(\bs{\beta}, h, \lambda_1, \gamma)$, 
which has a minimum value over the compact set $\|\bs{\beta}\|_2\leq \sqrt{t_2}$. 
\vs
{(ii)} Follows the approach originated in  Zuo and Zuo (2023) (ZZ23) or Zuo (2024), 
we partition the parameter space $\R^p$ of $\bs{\beta}$ into disjoint open pieces $R_{\bs{\beta}^k}$, $1\leq k\leq L$, 
$\cup_{1\leq k\leq L}\overline{R}_{\bs{\beta}^k}=\R^p$, where $\overline {A}$ stands for the closure of the set $A$, and
\be
R_{\bs{\beta}^k}=\{\bs{\beta}\in \R^p: I(\bs{\beta})=I(\bs{\beta}^k), r^2_{i_1}(\bs{\beta})< r^2_{i_2}(\bs{\beta})\cdots< r^2_{i_h}(\bs{\beta})\}, \label{region.eqn}
\ee
where 
$i_1,\cdots, i_{h}$ in $I(\bs{\beta})$ 
and 
with $w_i$ defined in (\ref{lts-en.eqn})
\be
I(\bs{\beta})=\Big\{ i:  w_i=1 
\Big\}. \label{I-beta.eqn}
\ee
For any $\bs{\beta} \in \R^p$, either there is $R_{\bs{\eta}}$ and
$\bs{\beta} \in R_{\bs{\eta}}$ or there is $R_{\bs{\xi}}$, such that $\bs{\beta} \not\in R_{\bs{\eta}}\cup R_{\bs{\xi}}$ and $\bs{\beta}\in \overline{R}_{\bs{\eta}}\cap \overline{R}_{\bs{\xi}}$. 
Now we claim that  $ \widehat{\bs{\beta}}:=\widehat{\bs{\beta}}^n_{lts-enet}(h, \lambda_1, \lambda_2, \gamma)\in R_{\bs{\beta}^{k_0}}$ for some $1\leq k_0\leq L$.
\vs
Otherwise, assume that
$\widehat{\bs{\beta}} \in \overline{R}_{\bs{\beta}^{k_0}}$. By Lemma 2.2 of ZZ23, $g(h, \bs{\beta}, \bs{Z}^{(n)})$ (denoted by $Q^{n}(\bs{\beta})$ there) is 
convex over 
 $R_{\bs{\beta}^{k_0}}$. Therefore, $O_n(h,\bs{\beta},\lambda_1, \lambda_2, \gamma)$ is strictly convex in $\bs{\beta}$ over  $R_{\bs{\beta}^{k_0}}$.  Assume that $\bs{\beta}^*$ is the global minimum of  $O_n(h,\bs{\beta},\lambda_1, \lambda_2, \gamma)$ over  $R_{\bs{\beta}^{k_0}}$. Then it is obviously that $O_n(h,\widehat{\bs{\beta}},\lambda_1, \lambda_2, \gamma)\leq O_n(h, \bs{\beta}^*,\lambda_1, \lambda_2, \gamma)$.
 But this is impossible in light of Lemma 3.2.
The strict convexity of $O_n(h,\bs{\beta},\lambda_1, \lambda_2, \gamma)$ over $R_{\bs{\beta}^{k_0}}$ guarantees the uniqueness.
\hfill \pend

\noin
 \tb{Proof of Lemma 2.2}
 \vs
 \noin
\tb{Proof:}  Let $B(\bs{x}^*, r)$ be a small open ball centered at $\bs{x}^*$ with a small radius $r$ and $  B(\bs{x}^*, r)\subset S$. Let $B^c:= S 
 -B(\bs{x}^*, r)$ and $\alpha^*=\inf_{\bs{x} \in B^c}f(\bs{x})$.
 Then, $\alpha^*>f(\bs{x}^*) $ (in light of strict convexity) 
 Since $y\in \overline{S}$, then there is a sequence $\{\bs{x}_j\} \in B^c$ such that $\bs{x}_j \to \bs{y}$ and $f(\bs{x}_j)\to f(\bs{y})$ (in light of continuity) as $j\to \infty$. Hence $f(\bs{y})=\lim_{j\to \infty}f(\bs{x}_j)\geq\alpha^*> f(\bs{x}^*)$. \hfill \pend
\vs
\vs\noin
\tb{Proof of Lemma 3.1}
\vs
\noindent
\tb{Proof}: Employing the facts that true model assumption: $\bs{Y}=\bs{X}\bs{\beta}_0+\bs{e}$ and that $\widehat{\bs{\beta}}^n$ is the minimizer of the RHS of (\ref{lts-en.eqn}), replacing $\bs{X}\bs{\beta}_0$ in the
LHS of (\ref{base-inequality.eqn}) with $\bs{Y}-\bs{e}$, expanding the LHS, leads to the desired result.
\hfill \pend
\vs
\noindent
\tb{Proof of Lemma 3.2}
\vs
\noindent
\tb{Proof}: (i) For any $D \in \mathbb{D}$,
let $i_1, \cdots, i_h$ be the sub-sequence from $\{1, \cdots, n\}$ such that the corresponding diagonal entries of $D$: $D(i_j, i_j)=1$ for $j \in \{1, \cdots, h\}$. It is readily seen that for any $s\in \R$
\begin{align}
&\mathbb{E}\Big(\exp\big(s(\bs{e}'D\bs{x}^{(j)}/\sqrt{n}\big)\Big) 
=\mathbb{E}\Big(\exp\big(s\frac{\sum_{k=1}^h \bs{e}_{i_k}x^{(j)}_{i_k}}{\sqrt{n}}\big)\Big)\nonumber\\[2ex]
&=\prod_{k=1}^h \mathbb{E}\exp\Big(\frac{s\bs{e}_{i_k}x^{(j)}_{i_k}}{\sqrt{n}}\Big)
\leq \prod_{k=1}^h \exp\Big(\frac{\sigma^2 s^2\big(x^{(j)}_{i_k}\big)^2/n}{2}\Big)\nonumber\\[2ex]
& =\exp\Big(\frac{s^2\sigma^2 \sum_{k=1}^h \big(x^{(j)}_{i_k}\big)^2/n}{2}\Big) \leq \exp\Big( \frac{s^2\sigma^2}{2}\Big).
\end{align}
The desired result (i) follows. \vs For (ii), notice that $\|\bs{e}\|^2_{D}/\sigma^2=\sum_{k=1}^h \bs{e}^2_{i_k}/\sigma^2$, i.e., 
the sum of h  squares of independent $N(0,1)$. 
 (ii) follows from the fact that $\bs{e}_{i_j}^2/\sigma^2$ has a $\chi^2$ distribution with one degree of freedom. 
 \vs
 (iii) The arguments above hold for any sub-sequence $i_1, \cdots, i_h$ from $\{1, \cdots, n\}$ uniformly. Equivalently, they hold true for any $D\in \mathbb{D}$ and  $\bs{\beta} \in \Theta$. 
 (iii) follows.
\hfill \pend
\vs
\noindent
\tb{Proof of Lemma 3.3}
\vs
\noin
\tb{Proof}: First we note that with $D=D_i$
\bee 
P\Big(\max_{1\leq j\leq p}|\bs{e}'D\bs{x}^{(j)}|/n>q_1/2\Big) 
&=& P\Big(\max_{1\leq j\leq p}\Big|\frac{\bs{e}'D\bs{x}^{(j)}}{\sqrt{n}}\Big| >\sqrt{n}q_1/2\Big)
\\[2ex]
&\leq& 2p \exp\big(\frac{-nq^2_1}{8\sigma^2}\big),
\ene
where the last inequality follows from the definition of the sub-Gaussian variable (see Theorem 2.2.1 of \cite{P20}) and the Lemma 3.2. Now set the right-hand side of the last display above to be $\delta/(2L)$, one gets the expression for $q_1$ and that
\bee
P\Big(\max_{1\leq j\leq p}|\bs{e}'D\bs{x}^{(j)}|/n>q_1/2\Big)
= P\Big(\max_{1\leq j\leq p}|\bs{e}'D\bs{x}^{(j)}/\sqrt{n}|>q_1\sqrt{n}/2\Big)\leq \delta/(2L),
\ene
the statement about $P(\mathscr{S}_1)$ in the Lemma follows. \vs
For the statement about $P(\mathscr{S}_2)$, we first invoke Lemma 4.2 and notice that $\|\bs{e}\|^2_{D(\bs{\beta}_0)}/\sigma^2$ follows a ${\chi}^2$ distribution with $h$ degrees of freedom,  then invoke Lemma 1 and Comments on page 1325 of \cite{LM00} 
 we get
$$
P\Big(\|\bs{e}\|^2_{D(\bs{\beta}_0)}/\sigma^2-h\geq 2\sqrt{ht}+2t \Big)\leq e^{-t},
$$
now if one sets $\delta/(4L)=e^{-t}$, that is $t=\log 4L/\delta$, then, restricted to $\Omega_i$, one gets 
$$P(\mathscr{S}_2)\geq p_i-\delta/(4L). $$
The second statement follows. Likewise, $\|\bs{e}\|^2_{D({\bs{\beta}})}/\sigma^2$ follows a ${\chi}^2$ distribution with $h$ degrees of freedom,  then invoke Lemma 1 and Comments on page 1325 of \cite{LM00},
 we get
$$
P\Big(h-\|\bs{e}\|^2_{D({\bs{\beta}})}/\sigma^2\geq 2\sqrt{ht} \Big)\leq e^{-t},
$$
Hence, restricted to $\Omega_i$,
$$P(\mathscr{S}_3)\geq p_i-\delta/(4L). $$
Statement on $P(\mathscr{S}_3)$ follows.
\vs 
Obviously, the exponential upper bounds above have nothing to do with any $D \in \mathbb{D}$ or any $D(\bs{\beta}) $ for any $\bs{\beta}\in \Theta$.  Hence, the probability inequalities in (\ref{probility.ineq}) hold uniformly over
$D\in \mathbb{D}$ and $D(\bs{\beta})$ with  any $\bs{\beta}\in \Theta$ and any $i \in \{1, 2, \cdots, L\}$.
 \hfill \pend 
  \vs\vs
\noindent
\tb{Proof of Lemma 3.4}
\vs
\noin 
\tb{Proof}: (i) For any $\bs{\beta}\in \Theta$, assume $D:=D(\bs{\beta})=D_i$ for some $i\in \{1,2,\cdots, L\}$,  
then 
\begin{align}
&\mathbb{G}(D, \bs{\beta}, \bs{e}; \bs{X}, \bs{\beta}_0,\lambda_1,\lambda_2)=\frac{2}{n}\bs{e}'D\bs{\beta}\bs{X}(\bs{\beta}-\bs{\beta}_0)+\frac{\sigma^2}{n}\left(\|\bs{e}\|^2_{D(\bs{\beta}_0)}/\sigma^2-\|\bs{e}\|^2_{D}/\sigma^2\right)\nonumber\\[2ex]
&+g(\bs{\beta}, \bs{\beta}_0,\lambda_1, \lambda_2) \nonumber 
\nonumber\\[2ex]
&\leq \frac{2}{n}\Big|\max_{1\leq j\leq p}\bs{e}'D\bs{x^{(j)}}\Big|\Big(\|\bs{\beta}-\bs{\beta}_0\|_{1}\Big)+ \frac{\sigma^2}{n}\left(\|\bs{e}\|^2_{D(\bs{\beta}_0)}/\sigma^2-\|\bs{e}\|^2_{D}/\sigma^2\right)+g(\bs{\beta}, \bs{\beta}_0,\lambda_1, \lambda_2),\nonumber\\[2ex]
&\leq \frac{2}{n}\Big|\max_{1\leq j\leq p}\bs{e}'D\bs{x^{(j)}}\Big|\Big(\|\bs{\beta}\|_1+\|\bs{\beta}_0\|_{1}\Big)+ \frac{\sigma^2}{n}\left(\|\bs{e}\|^2_{D(\bs{\beta}_0)}/\sigma^2-\|\bs{e}\|^2_{D}/\sigma^2\right)+g(\bs{\beta}, \bs{\beta}_0,\lambda_1, \lambda_2),\nonumber\\[2ex]
&\leq \|\bs{\beta}\|_1\left(\frac{2}{n}\Big|\max_{1\leq j\leq p}\bs{e}'D\bs{x^{(j)}}\Big|-\lambda_1\right)+\|\bs{\beta}_0\|_1\left(\frac{2}{n}\Big|\max_{1\leq j\leq p}\bs{e}'D\bs{x^{(j)}}\Big|+\lambda_1(1+\|\bs{\beta}_0\|_1)\right)\nonumber\\[1ex]
&+\frac{\sigma^2}{n}\left(\|\bs{e}\|^2_{D(\bs{\beta}_0)}/\sigma^2-\|\bs{e}\|^2_{D}/\sigma^2\right), \label{new-lemma4.4.eqn}
\end{align}
where  the facts that $\lambda_2\leq \lambda_1$ and $\|x\|_2\leq \|x\|_1$ are employed.
\vs
In light of Lemma 3.3,  we have that
\bee
P\left(\frac{2}{n}\Big|\max_{1\leq j\leq p}\bs{e}'D\bs{x^{(j)}}\Big| >q_1\right)&\leq&
\delta/(2L)\\[1ex]
P\left(\|\bs{e}\|^2_{D(\bs{\beta}_0)}/\sigma^2-h>q_2 \right)&\leq& \delta/(4L);~~
P\left(h-\|\bs{e}\|^2_{D}/\sigma^2>q_3 \right)\leq \delta/(4L).
\ene
These combined with (\ref{new-lemma4.4.eqn}) imply that for any $\bs{\beta} \in \Theta$ with $D(\bs{\beta})=D_i$ for some $i \in \{1,2,\cdots, L\}$, restricted to $\Omega_i$ 
with a probability of at least 
$p_i-\delta/L$ we have that
\begin{align}
\mathbb{G}(D(\bs{\beta}), \bs{\beta}, \bs{e}; \bs{X}, \bs{\beta}_0,\lambda_1,\lambda_2) 
&\leq \|\bs{\beta}_0\|_1(\lambda_1+\lambda_1\big(1+\|\bs{\beta}_0\|_1)\big) +\frac{\sigma^2}{n}(q_2+q_3)\nonumber\\[2ex]
&\leq\lambda_1 \|\bs{\beta}_0\|_1 (2+\|\bs{\beta}_0\|_1) + \|\bs{\beta}_0\|_1 (2 q_2)\frac{\sigma^2}{n}  /\|\bs{\beta}_0\|_1 \nonumber\\[1ex]
&= \frac{\sigma^2}{n} \|\bs{\beta}_0\|_1 C(\bs{\beta}_0,n, \sigma,q_1, q_2),  \label{upperbound.ineq}
\end{align}
where $C(\bs{\beta}_0,n, \sigma,q_1, q_2)=2nq_1(2+\|\bs{\beta}_0\|_1)/\sigma^2+2q_2/\|\bs{\beta}_0\|_1:=Const$. That is, 

\bee
&& P\left(\omega \in \Omega_i, D(\bs{\beta})=D_i, \bs{\beta}\in \Theta: \mathbb{G}(D(\bs{\beta}), \bs{\beta}, \bs{e}; \bs{X}, \bs{\beta}_0,\lambda_1,\lambda_2)\leq \frac{\sigma^2}{n} \|\bs{\beta}_0\|_1 Const \right)\\[2ex] &&\geq  p_i-\delta/L.
\ene

\vs
\vs
(ii) Write the upper bound in (\ref{upperbound.ineq}) as UPBD. Over the entire $\Omega$, one has in light of (i) above
\begin{align}
&P\Big(\omega \in \Omega,  \bs{\beta}\in \Theta: \mathbb{G}(D(\bs{\beta}), \bs{\beta}, \bs{e}; \bs{X}, \bs{\beta}_0,\lambda_1,\lambda_2)\leq \mbox{UPBD} \Big)\nonumber \\[2ex]
&=
\sum_{i=1}^L P\Big(\omega \in \Omega_i, D(\bs{\beta})=D_i, \bs{\beta}\in \Theta: \mathbb{G}(D(\bs{\beta}), \bs{\beta}, \bs{e}; \bs{X}, \bs{\beta}_0,\lambda_1,\lambda_2)\leq \mbox{UPBD} \Big)
 \nonumber \\[2ex]
 &\geq \sum_{i=1}^L (p_i-\delta/L)=1-\delta.
\end{align}
\hfill \pend
 \vs
\noindent
\tb{Proof of Theorem 3.1}
\vs
\noin 
\tb{Proof}: In light of Lemma 3.4, one has with probability at least $1-\delta$ for any $\delta \in (0,1)$,
$$
\mathbb{G}(D(\bs{\beta}), \bs{\beta}, \bs{e}; \bs{X}, \bs{\beta}_0,\lambda_1,\lambda_2)
\leq \frac{\sigma^2}{n} \|\bs{\beta}_0\|_1 C(\bs{\beta}_0,n, \sigma,q_1, q_2). $$
Write the RHS above as UPBD, then by Lemma 3.1,
one has \begin{align*}P\left(\frac{1}{n}\|\bs{X}(\widehat{\bs{\beta}}^n-\bs{\beta}_0\|^2_{D(\widehat{\bs{\beta}}^n)}\leq \mbox{UPBD}\right) &\geq P\Big(\mathbb{G}(D(\widehat{\bs{\beta}}^n), \widehat{\bs{\beta}}^n, \bs{e}; \bs{X}, \bs{\beta}_0,\lambda_1,\lambda_2)\leq \mbox{UPBD}\Big)\\[2ex]
&\geq 1-\delta.\end{align*}
The desired result follows.
\hfill \pend 
 \vs
\noindent
\tb{Proof of Lemma 3.5}
\vs
\noin 
\tb{Proof}: In light of Lemma 3.1, one can write
\begin{align*}
&\frac{2}{n}\|\bs{X}(\widehat{\bs{\beta}}^n-\bs{\beta}^0)\|^2_{D(\widehat{\bs{\beta}}^n)}
+2\lambda\|\widehat{\bs{\beta}}^n\|_1 \\[1ex]
&\leq  \frac{4}{n}\bs{e}'D(\widehat{\bs{\beta}}^n)\bs{X}(\widehat{\bs{\beta}}^n-\bs{\beta}^0)+2\lambda\|\bs{\beta}^0\|_1+\frac{2\sigma^2}{n}\left(\|\bs{e}\|^2_{D(\bs{\beta}_0)}/\sigma^2-\|\bs{e}\|^2_{D(\widehat{\bs{\beta}}^n)}/\sigma^2\right) \\[1ex]
&\leq 2\Big|\max_{1\leq j\leq p}\frac{2}{n}\bs{e}'D(\widehat{\bs{\beta}}^n)\bs{x}^{(j)}\Big|\|\widehat{\bs{\beta}}^n-\bs{\beta}^0\|_{1}+2\lambda\|\bs{\beta}^0\|_1+\frac{2\sigma^2}{n}\left(\|\bs{e}\|^2_{D(\bs{\beta}_0)}/\sigma^2-\|\bs{e}\|^2_{D(\widehat{\bs{\beta}}^n)}/\sigma^2\right)
\end{align*}
In light of Lemma 3.3 and (ii) of Remarks 3.1, with probability at least $1-\delta$, we have
\begin{align}
&\frac{2}{n}\|\bs{X}(\widehat{\bs{\beta}}^n-\bs{\beta}^0)\|^2_{D(\widehat{\bs{\beta}}^n)}
+2\lambda\|\widehat{\bs{\beta}}^n\|_1 \nonumber\\[1ex]
&\leq \lambda\|\widehat{\bs{\beta}}^n-\bs{\beta}^0\|_{1}+2\lambda\|\bs{\beta}^0\|_1+\widehat{\sigma}^2\frac{4q_2}{n}. \label{lemma4.4}
\end{align}
For notation simplicity, write $\hat{\bs{\beta}}$ for $\widehat{\bs{\beta}}^n$, we have
\begin{align*}
&\|\hat{\bs{\beta}}^n\|_1= \|\hat{\bs{\beta}}_{S_0}\|_1+\|\hat{\bs{\beta}}_{S^c_0}\|_1\geq
\|\hat{\bs{\beta}}_{S^c_0}\|_1+\|\bs{\beta}^0_{S_0}\|_1-\|\hat{\bs{\beta}}_{S_0}-\bs{\beta}^0_{S_0}\|_1\\[1ex]
&\|\widehat{\bs{\beta}}^n-\bs{\beta}^0\|_{1} \leq \|\hat{\bs{\beta}}_{S_0}-\bs{\beta}^0_{S_0}\|_1+\|\hat{\bs{\beta}}_{S^c_0}\|_1
\end{align*}
The first inequality above leads to 
\[
\|\bs{\beta}_0\|_1\leq \|\hat{\bs{\beta}}_{S_0}-\bs{\beta}^0_{S_0}\|_1+\|\hat{\bs{\beta}}_{S_0}\|_1,
\]
which, combining with the second inequality above, leads to
\[
\lambda\|\widehat{\bs{\beta}}^n-\bs{\beta}^0\|_{1}+2\lambda\|\bs{\beta}^0\|_1\leq 3\lambda\|\hat{\bs{\beta}}_{S_0}-\bs{\beta}^0_{S_0}\|_1-\lambda\|\hat{\bs{\beta}}_{S^c_0}\|_1 + 2\lambda\|\widehat{\bs{\beta}}^n\|_1
\]
This, in conjunction with (\ref{lemma4.4}) and the definition of $\eta$, leads to the desired result. \hfill \pend
\vs
\noin
\tb{Proof of Theorem 3.2}
\vs
\noin 
\tb{Proof}: For convenience, write $\theta$ for $\widehat{\bs{\beta}}^n-\bs{\beta}^0$, then it is seen that
\[\mbox{MSE}(\bs{X}_*\widehat{\bs{\beta}}^n)=\theta^{\top}\frac{\bs{X}^{\top}_*\bs{X}_*}{n}\theta=\|\bs{X}_*\theta\|^2_2/n=\frac{1}{n}\Big(\|\bs{X}_*\theta_{S_0}\|^2_2+\|\bs{X}_*\theta_{S^c_0}\|^2_2)+2\theta^{\top}_{S_0}\bs{X}^{\top}_*\bs{X}_*\theta_{S^c_0}\Big).
\] 
Now we treat the three terms on the right-hand side above.
\[
\|\bs{X}_*\theta_{S_0}\|^2_2=n\|\theta_{S_0}\|^2_2+n\theta^{\top}_{S_0}\big(\frac{\bs{X}^{\top}_*\bs{X}_*}{n}-\mathbb{I}_{p \times p}\big)\theta_{S_0}\geq n\|\theta_{S_0}\|^2_2-n \frac{\|\theta_{S_0}\|^2_1}{32k},
\]
in light of the incoherence assumption and the H\"{o}lder's inequality, note that $D(\widehat{\bs{\beta}}^n)$ belongs to $\mathbb{D}$.
Likewise
\begin{align*}
\|\bs{X}_*\theta_{S^c_0}\|^2_2&=n\|\theta_{S^c_0}\|^2_2+n\theta^{\top}_{S^c_0}\big(\frac{\bs{X}^{\top}_*\bs{X}_*}{n}-\mathbb{I}_{p \times p}\big)\theta_{S^c_0}\\[1ex]
&\geq n\|\theta_{S^c_0}\|^2_2-n \frac{\|\theta_{S^c_0}\|^2_1}{32k}\\[1ex]
&\geq n\|\theta_{S^c_0}\|^2_2-n \frac{(3\|\theta_{S_0}\|_1+\zeta)^2}{32k}\\[1ex]
&\geq n\|\theta_{S^c_0}\|^2_2-\frac{2n\Big(9\|\theta_{S_0}\|^2_1+\zeta^2\Big)}{32k},
\end{align*}
in light of Lemma 3.5 and $(a+b)^2\leq 2(a^2+b^2)$, where $\zeta:=\eta/\lambda$.
Finally, employing the incoherence and the H\"{o}lder's inequality and the definition of $\bs{\beta}_S$, for any $\bs{\beta}\in \R^p$ and index set $S$, one has
\[2\big|\theta^{\top}_{S_0}\bs{X}^{\top}_*\bs{X}_*\theta_{S^c_0}\big|\leq \frac{2n}{32k}\|\theta_{S_0}\|_1\|\theta_{S^c_0}\|_1\leq \frac{n}{32k}\left(\|\theta_{S_0}\|^2_1 +\|\theta_{S^c_0}\|^2_1\right),
\]
in light of $2ab\leq a^2+b^2$. Combining the three results for the three terms above, one has
\begin{align}
\|\bs{X}_*\theta\|^2_2/n&\geq \|\theta_{S_0}\|^2_2+\|\theta_{S^c_0}\|^2_2-\frac{1}{32k}
\left(20\|\theta_{S_0}\|^2_1+\|\theta_{S^c_0}\|^2_1 +2\zeta^2\right)\nonumber\\[2ex]
&\geq \|\theta_{S_0}\|^2_2+\|\theta_{S^c_0}\|^2_2 -\frac{20}{32} \|\theta_{S_0}\|^2_2 -\frac{p-s_0}{32k}\|\theta_{S^c_0}\|^2_2 -\frac{2}{32k}\zeta^2\nonumber \\[2ex]
&=\frac{3\|\theta_{S_0}\|^2_2}{8} +\Big(1-\frac{p-s_0}{32k}\Big)\|\theta_{S^c_0}\|^2_2-\frac{2}{32k}\zeta^2
 \nonumber \\[1ex]
&\geq \frac{3}{8}\|\theta\|^2_2-\frac{1}{16k}\zeta^2,\nonumber
\end{align}
where the facts that $(|a_1|+\cdots+|a_n|)^2\leq n(a_1^2+\cdots+a_n^2)$ (i.e.  H\"{o}lder's inequality) and $k\geq
\min\{s_0, (p-s_0)/20\}$ are utilized.
The desired result follows.
\hfill \pend
\vs

\end{document}